\pdfoutput=1
\documentclass[a4paper]{article}

\usepackage{INTERSPEECH2019}
\usepackage{color}

\title{Learning pronunciation from a foreign language in speech synthesis networks}
\name{Younggun Lee$^1$, Suwon Shon$^2$, Taesu Kim$^1$}
\address{
  $^1$Neosapience, Inc., Seoul, South Korea\\
  $^2$Massachusetts Institute of Technology, Cambridge, MA, USA}
\email{\{yg,taesu\}@neosapience.com, swshon@mit.edu}

\begin{document}

\maketitle
\begin{abstract}
Although there are more than 6,500 languages in the world, the pronunciations of many phonemes sound similar across the languages. When people learn a foreign language, their pronunciation often reflects their native language's characteristics. This motivates us to investigate how the speech synthesis network learns the pronunciation from datasets from different languages. In this study, we are interested in analyzing and taking advantage of multilingual speech synthesis network. First, we train the speech synthesis network bilingually in English and Korean and analyze how the network learns the relations of phoneme pronunciation between the languages. Our experimental result shows that the learned phoneme embedding vectors are located closer if their pronunciations are similar across the languages. Consequently, the trained networks can synthesize the English speakers' Korean speech and vice versa. Using this result, we propose a training framework to utilize information from a different language. To be specific, we pre-train a speech synthesis network using datasets from both high-resource language and low-resource language, then we fine-tune the network using the low-resource language dataset. Finally, we conducted more simulations on 10 different languages to show it is generally extendable to other languages.
\end{abstract}

\section{Introduction}
Among many languages in the world, some of the languages have phonemes with the same pronunciation. Conceptually, if the intersection of the pronunciations of two languages is large enough, the burden of learning the pronunciation of the one language after acquiring the other language will be lowered. We can also see people tends to show the accent of their first language when speaking in their second language \cite{2ndlang}. We can also recognize easily that people tend to use their native language accent when speaking a foreign language. This may tell us that people learn new pronunciation based on their native language. Motivated by this, we investigated how the datasets from different languages can help training and generalization of speech synthesis model.

At first, we trained a bilingual TTS model using monolingual speakers' speech database. Deep neural network based Text-To-Speech (TTS) models allow us to build TTS models that can generate natural-sounding speech \cite{Tacotron, DV, DV2, TransformerTTS}. Among the various approaches, end-to-end TTS models, such as Tacotron, require us only a little amount of prior knowledge about languages, and the TTS model can be trained easily if we have enough amount of text-speech pair data. Also, given enough amount of multiple speakers' speech data, we can build a multi-speaker TTS model when the amount of each speakers' speech data is relatively small \cite{DV2}. In a similar way, we trained a multilingual multi-speaker TTS model using text-speech pairs of English and Korean. 
Interestingly, although we did not have any speaker who speaks both English and Korean, every speaker in the trained model could speak both English and Korean. To investigate how the model can generate a foreign language, we extracted phoneme embeddings and measured similarities between English and Korean phonemes. We found phonemes that have similar pronunciation tend to stay closer than the others even across the different languages. This result shows us that the interaction between two phonetic systems could help to generate foreign language using only the native language.

Based on this phonetic system interaction behavior, we hypothesized that the multilingual TTS model can be useful for training a TTS model of low-resource language. The major obstacles to train a TTS model for low-resource language is that lack of linguistic analysis components, such as dependency tree parser and morphological analyzer, and data itself \cite{Goog17}. Without linguistic analysis component, we cannot obtain linguistic features to train a TTS model. This problem can be circumvented with the help of end-to-end neural TTS models which require minimal prior knowledge. On the other hand, TTS models usually need a large amount of text-speech pair data for training. According to \cite{Data_amount}, original Tacotron requires more than 10 hours of speech data from one speaker to obtain good generation quality. It is reported that the model can benefit from pre-training Tacotron decoder with a large amount of speech data \cite{Data_amount}, but such large data is rarely available for the low-resource languages. Recall that, the learned phoneme embedding in multilingual TTS model tends to represent phoneme's pronunciation. Combining this observation with the pre-training approach led us to pre-train TTS model using a large amount of text-speech pair data from a high-resource language together with a small amount of text-speech pair from a low-resource language. The pre-trained model demonstrated higher performance in both subjective and objective measure. Moreover, we used the proposed approach to ten different languages to see if this method works for the other combination of languages.

To summarize, the contributions of this study are as follows:
\begin{enumerate}
  \item We find two learned phoneme embeddings from different languages are located close if the pronunciation of the two phonemes is similar.
  \item We show that pre-training a model with a large amount of data from high-resource language and target speaker's data can enhance the performance of the TTS model in low-resource language.
  \item We validate the proposed pre-training framework by applying it to training TTS models of ten different languages.
\end{enumerate}

\section{Proposed method}
\label{method}

\subsection{Multilingual multi-speaker Tacotron}
\label{model}
We implement a multilingual multi-speaker TTS model by modifying Tacotron \cite{Tacotron}. We use simplified version \cite{Tacotron2} of Tacotron for the encoder and the decoder, but we use the original Tacotron style of Post-processing net and Griffin-Lim algorithm \cite{GLRecon} for conversion of a linear-scale spectrogram to a waveform. A sequence of phonemes are converted to phoneme embeddings, then fed to the encoder as input. We concatenate phoneme embedding dictionary of each language to form the entire phoneme embedding dictionary, so there may exist duplicated phonemes in the dictionary if the languages share same phonemes. Note that, the phoneme embeddings are normalized to have the same norm. In order to model multiple speakers' voices in a single TTS model, we adopt Deep Voice 2 \cite{DV2} style speaker embedding network. One-hot speaker identity vector is converted to a 32-dimensional speaker embedding vector by the speaker embedding network. Unless stated otherwise, we use the same hyperparameter settings with \cite{Tacotron2}.

The model is trained to minimize L1 losses between ground-truth spectrogram and predicted spectrogram for both the linear-scale spectrogram and the Mel-scale spectrogram. When we train a multi-speaker TTS model, the amount of speech data differs across speakers. This data imbalance may induce a bias to the TTS model. To cope with this data imbalance, we divide the loss of each sample from one speaker by the total number of samples in a training set which belongs to the speaker. We empirically found that this adjustment in loss function yields better synthesis quality.

\subsection{Multilingual pre-training}
\label{pre-train}
In the analysis of Subsection \ref{exp-phoneme}, having observed the learned phoneme embeddings from multilingual multi-speaker Tacotron, we noticed that two learned phoneme embeddings are located closer if they have similar pronunciation. We think that learning to generate speech in one language will help to learn to generate speech in another language because some information can be shared across the languages, such as phoneme-pronunciation relation of phonemes exist in common. We implemented this idea by pre-training a multilingual multi-speaker Tacotron with datasets from two languages.

We use the multilingual multi-speaker Tacotron for multilingual pre-training. Given a small amount of low-resource language text-speech pairs and a large amount of high-resource language text-speech pairs, we use the union of them to pre-train the model. There can be many ways to utilize information from other languages, such as decoder pre-training \cite{Data_amount}. However, we chose to pre-train the entire model since it does not incur mismatch between pre-train and fine-tune as described in Section \ref{related}. Also, this model can potentially benefit from phoneme-pronunciation relation of languages learned in pre-training while decoder pre-training approach cannot use this information. 

The pre-training is ceased when the validation loss stops to decrease. After that, the model is fine-tuned with the same text-speech pairs of the low-resource language. 

\section{Related work}
\label{related}
For multilingual speech synthesis, some researchers used International Phonetic Alphabet (IPA), which can cover phones of all training languages, to transcribe texts of multilingual dataset \cite{IEEE12, IPA}. Alternative approaches other than using IPA were reported by taking a union of multiple languages' linguistic features \cite{Goog16} or defining new linguistic features \cite{Goog17}. However, these approaches still require prior knowledge of the low-resource languages to obtain linguistic features. End-to-end neural TTS models can lower the barrier of research and development of TTS models for low-resource languages since it requires only text-speech pairs. Also, additional information, such as grapheme-to-phoneme relation, can be easily integrated into the end-to-end neural TTS model to improve synthesis quality by replacing grapheme sequences with phoneme sequences. 

Chung et al. proposed a data efficient semi-supervised training of TTS model that trains the model using a small amount of text-speech pair data with help of decoder pre-training \cite{Data_amount}. They pre-trained decoder of Tacotron using only speech data while substituting the context vector with zero vector. From the pre-training, the decoder can learn how to generate acoustic representations, say Mel-spectrogram, in an auto-regressive way. After the pre-training, they fine-tuned the network with the target speaker's dataset. One shortcoming of this decoder pre-training is a mismatch in the distribution of the context vector between the training phase and the test phase. Also, the proposed pre-training approach inevitably assumes the existence of a large amount of speech data which may not be available for low-resource languages. Compared to this approach, our proposed method has lesser mismatch by using phoneme input during the pre-training, and it does not require large data for low-resource languages. Furthermore, our approach uses the target speaker's dataset at pre-train as well as at fine-tune. Since the target speaker's data is accessible at pre-training, there is no reason to avoid using it. It may increase the training time, but we think an improvement in generation quality by using the target speaker's data is more valuable than saving training time.

Other researchers were interested in transferring voice across languages \cite{JiaIMI}, yet they only focused on transferring voice identity, not pronunciations of languages. 

\section{Experiments and results}
\label{exp}
\subsection{Dataset}
In Subsection \ref{exp-phoneme}, we used 60 hours of English data and 60 hours of Korean data. For English dataset, we used the union of VCTK dataset, CMU Arctic, and LJSpeech \cite{vctk, cmuarctic, ljspeech}. For Korean dataset, we used our proprietary multi-speaker dataset. In Subsection \ref{exp-pretrain}, we used the same Korean dataset and English dataset for pre-training and 2013 Blizzard Challenge dataset for an English target speaker's dataset. In Subsection \ref{exp-css}, we additionally used CSS10 dataset which is the collection of single speaker speech datasets for ten different languages \cite{css10}. The ten languages include Chinese, Dutch, Finnish, French, German, Greek, Hungarian, Japanese, Russian, and Spanish.

We used grapheme-to-phoneme (G2P) libraries to convert text to the corresponding phoneme sequence. For English text, we used a G2P library \cite{eng2p} which extends CMUdict \cite{CMUdict} phoneme dictionary by predicting phonemes of out-of-vocabulary words with a neural network. For Korean text, we used KoG2P \cite{kog2p} which is commonly used for Korean text preprocessing. We did not use G2P for CSS10 dataset, and we used pinyin transcript and romanized transcript for Chinese and Japanese respectively.

\subsection{Training a bilingual multi-speaker TTS}
\label{exp-phoneme}
We investigated how the phoneme representation is learned when the network is trained with the dataset containing both English and Korean.

It is known that the pronunciation of different languages can be described by one unified alphabetic system, IPA. The conversion tables of IPA-CMUdict and IPA-KoG2P can be found in \cite{ENG-IPA} and appendix respectively. Although one-to-one correspondence does not hold between English phoneme set and Korean phoneme set in terms of IPA, we carefully chose a subset of each set to include only the phonemes that have the common pronunciations in both languages. The chosen subsets can be found in the appendix.

For each phoneme in the chosen subsets, anchor phoneme, we computed cosine distance between each of the other phonemes. Then, we listed the 4-nearest phoneme embedding of the opposite language to analyze the learned phoneme representation. The results are shown in Table \ref{table-nearest-ko} and Table \ref{table-nearest-en}. The numbers 0, 1, and 2 after the English phonemes denote "No stress", "Primary stress", and "Secondary stress" respectively. While CMUdict distinguished stressed pronunciation, we could not find corresponding IPA representation, so we reported all stress types. The results show that most of the anchor phonemes' corresponding phonemes in the opposite language were found in the 4-nearest phoneme embeddings. As shown in Table \ref{table-nearest-en}, the 4-nearest phonemes of 70 phonemes (57 phonemes in Korean) include the corresponding phonemes and similar pronunciations. It implies that the phoneme embeddings learned the relation of pronunciations across the languages.

\begin{table}[t]
  \caption{The 4-nearest English phonemes of each Korean phoneme whose pronunciation exists in English, IPA symbols are written in the parentheses}
  \label{table-nearest-ko}
  \centering
  \begin{tabular}{ccccc}
    \toprule
    Korean\\
    phoneme & 1st & 2nd & 3rd & 4th\\
    \cmidrule(r){1-5}
h0 (h) & \hl{HH (h)} & AH0 (\textipa{2}) & AE0 (\ae) & UW (u)\\			
ii (i) & \hl{IY2 (i)} & Y (j) & UW2 (u) & UW1 (u)\\			
k0 (g) & \hl{G (g)} & K (k) & DH (\dh) & V (v)\\			
kf (k) & \hl{K (k)} & UW (u) & CH (t\textipa{S}) & G (g)\\			
ll (l) & R (\textturnr) & ER1 (\textrhookrevepsilon) & ER2 (\textrhookrevepsilon) & Y (j)\\			
ng (\textipa{N}) & M (m) & N (n) & \hl{NG (\textipa{N})} & UH2 (\textipa{U})\\			
p0 (b) & \hl{B (b)} & OY0 (\textipa{OI}) & V (v) & P (p)\\			
pf (p) & UW (u) & B (b) & \hl{P (p)} & OY0 (\textipa{OI})\\			
qq (\textipa{E}) & EY1 (e\textipa{I}) & \hl{EH1 (\textipa{E})} & EY0 (e\textipa{I}) & ER1 (\textrhookrevepsilon)\\
rr (\textipa{R}) & IH0 (\textipa{I}) & AE0 (\ae) & UW (u) & AH0 (\textipa{2})\\			
t0 (d) & \hl{D (d)} & T (t) & TH (\textipa{T}) & G (g)\\			
tf (t) & UW (u) & \hl{T (t)} & AE2 (\ae) & D (d)\\			
uu (u) & \hl{UW0 (u)} & OY2 (\textipa{OI}) & OY1 (\textipa{OI}) & W (w)\\			
vv (\textipa{2}) & AO0 (\textipa{O}) & AA0 (\textipa{A}) & \hl{AH2 (\textipa{2})} & AA2 (\textipa{A})\\			
    \bottomrule
  \end{tabular}
\end{table}

\begin{table}[t]
  \caption{The 4-nearest Korean phonemes of each English phoneme whose pronunciation exists in Korean, IPA symbols are written in the parentheses, '(-)' denotes that there is no exact IPA symbol for that phoneme.}
  \label{table-nearest-en}
  \centering
  \begin{tabular}{cccccc}
    \toprule
    English\\
    phoneme & 1st & 2nd & 3rd & 4th\\
    \cmidrule(r){1-5}
AH0 (\textipa{2}) & xx (\textipa{W}) & rr (\textipa{R}) & ps (-) & \hl{vv (\textipa{2})}\\
AH1 (\textipa{2}) & \hl{vv (\textipa{2})} & ls (-) & wv (w) & wa (w)\\
AH2 (\textipa{2}) & \hl{vv (\textipa{2})} & lk (-) & lb (-) & aa (a)\\
B (b) & pp (\textipa{\"*p}) & kk (\textipa{\"*k}) & \hl{p0 (b)} & tt (\textipa{\"*t})\\
D (d) & tt (\textipa{\"*t}) & \hl{t0 (d)} & cc (t\textipa{\"*s}) & c0 (dz)\\
EH0 (\textipa{E}) & ls (-) & lh (-) & ye (j) & aa (a)\\
EH1 (\textipa{E}) & ya (j) & ee (e) & \hl{qq (\textipa{E})} & aa (a)\\
EH2 (\textipa{E}) & lm (-) & ya (j) & rr (\textipa{R}) & \hl{qq (\textipa{E})}\\
G (g) & kk (\textipa{\"*k}) & \hl{k0 (g)} & tt (\textipa{\"*t}) & yq (j)\\
IY0 (i) & wi (w) & xi (\textipa{W}i) & ee (e) & ls (-)\\
IY1 (i) & wi (w) & yq (j) & xi (\textipa{W}i) & \hl{ii (i)}\\
IY2 (i) & we (w) & \hl{ii (i)} & wi (w) & ye (j)\\
K (k) & kh (\textipa{k\super{h}}) & kk (\textipa{\"*k}) & \hl{kf (k)} & tt (\textipa{\"*t})\\
L (l) & vv (\textipa{2}) & uu (u) & wv (w) & rr (\textipa{R})\\
M (m) & \hl{mf (m)} & \hl{mm (m)} & ng (\textipa{N}) & nf (n)\\
N (n) & \hl{nn (n)} & \hl{nf (n)} & ng (\textipa{N}) & mm (m)\\
P (p) & ph (\textipa{p\super{h}}) & pp (\textipa{\"*p}) & kh (\textipa{k\super{h}}) & tt (\textipa{\"*t})\\
T (t) & th (\textipa{t\super{h}}) & ch (t\textipa{s\super{h}}) & tt (\textipa{\"*t}) & ph (\textipa{p\super{h}})\\
UW (u) & pf (p) & tf (t) & rr (\textipa{R}) & kf (k)\\
UW0 (u) & yu (-) & \hl{uu (u)} & yo (-) & wv (w)\\
UW1 (u) & yu (-) & wi (w) & \hl{uu (u)} & ii (i)\\
UW2 (u) & yo (-) & yu (-) & we (w) & wq (w)\\
W (w) & oo (o) & pf (p) & uu (u) & kf (k)\\
Y (j) & \hl{yq (j)} & ii (i) & \hl{yu (-)} & ll (l)\\
    \bottomrule
  \end{tabular}
\end{table}

Although not all of them have the corresponding pronunciation in the other language, pronunciation of the nearest phonemes sounded similar pronunciation. To check their perceptual similarity, we first generated speech of an English sentence, and we replaced each phoneme with the nearest Korean phoneme. Readers can find that both of the generated speeches sounds similarly in our demo page.\footnote{http://neosapience.com/research/demo-learning-pron/}

\subsection{Improved TTS model by multilingual pre-training}
In this experiment, we hypothetically assumed English as a low-resource language since it is easier for readers to compare the quality of synthesized speech in English than other languages. We trained an English TTS model given a small amount of English dataset from a target speaker (Low-resource) and a large amount of Korean dataset (High-resource). Note that, we are not accessible to other English datasets since we are assuming English as a low-resource language. We first pre-trained TTS models to use information of other language and then fine-tuned them with low-resource target speaker data. There are three possible combinations of data for pre-training: a low-resource language only (equivalent with no pre-train), a high-resource language only, and union of them. We denoted the three ways of pre-train as T-base, PD-H, and PA-HL; details are explained in the following context.

The baseline, T-base, was not pre-trained at all. For PD-H (Pre-train Decoder with High-resource language), only the decoder module was pre-trained with 60 hours of Korean speech data following the method of \cite{Data_amount}. PA-HL (Pre-train All modules with High- and Low-resource language) was pre-trained with 60 hours of Korean text-speech pairs and the small English text-speech pairs of the target-speaker as described in Subsection \ref{pre-train}. In addition to the three models, we compared a model PD-E which is pre-trained in the same way as PD-H except that we used 60 hours of English speech data instead of the Korean speech data. Note that, we may not be able to obtain PD-E in practice, since it can be difficult to find a large speech dataset for low-resource language. All of them have the same architecture described in Subsection \ref{model}. 

After the pre-training, we fine-tuned each model with the English dataset of the target speaker. In the fine-tuning phase, we varied amounts of the target speaker's data used by 0.4, 2, and 10 hours. The fine-tuned models were compared through side-by-side preference test and calculating word error rate (WER). For the evaluation, we generated speech samples using 100 unseen test sentences. The test sentences were randomly selected from Harvard sentences \cite{harvard}. We crowd-sourced workers using Amazon Mechanical Turk to compare speech samples from two different models. When calculating WER, we first fed generated speech samples to Google speech recognition API and calculated WER using the recognition result and the ground truth sentence. The result of the preference test and WER are summarized in Table \ref{table-preference-bli} and Table \ref{table-wer} respectively, and the generated samples from each model are posted in the demo page.
\label{exp-pretrain}

\begin{table}[t]
  \caption{Result of preference test based on a 7-point rating scale.}
  \label{table-preference-bli}
  \centering
  \begin{tabular}{cccccc}
    \toprule
    \multirow{3}{*}{\shortstack[c]{Data\\(hr)}} & \multirow{3}{*}{\shortstack[c]{Competing\\pair}} & \multicolumn{3}{c}{Preference ($\%$)} \\
    \cmidrule(r){3-5}
     & & Former & Neutral & Latter \\
    \cmidrule(r){1-5}
    \multirow{1}{*}{\shortstack[c]{0.4}}  & PD-E vs. PA-HL & 29.7 & 16.3 & \textbf{54.0}  \\
    \cmidrule(r){1-5}
    \multirow{2}{*}{\shortstack[c]{2}}  & PD-H vs. PA-HL & 25.0 & 25.3 & \textbf{49.7}  \\
     & PD-E vs. PA-HL & 31.0 & 24.7 & \textbf{44.3} \\
    \cmidrule(r){1-5}
    \multirow{3}{*}{\shortstack[c]{10}}  & T-ori vs. PA-HL & 20.3 & 11.3 & \textbf{68.3}  \\
     & PD-H vs. PA-HL & 26.0 & 29.3 & \textbf{44.7} \\
     & PD-E vs. PA-HL & 33.0 & 22.7 & \textbf{44.3} \\
    \bottomrule
  \end{tabular}
\end{table}

\begin{table}[t]
  \caption{Word error rate (\%) by types of pre-training and amounts of data used for fine-tune.}
  \label{table-wer}
  \centering
  \begin{tabular}{ccccc}
    \toprule
    \multirow{3}{*}{\shortstack[c]{Model}} & \multicolumn{3}{c}{Fine-tune data size (hr)} \\
    \cmidrule(r){2-4}
     & 0.4 & 2 & 10 \\
    \cmidrule(r){1-4}
    T-base & n/a & n/a & 26.3 \\
    PD-H & n/a & 37.4 & 21.1 \\
    PA-HL & 55.8 & \textbf{30.1} & \textbf{15.0} \\
    \cmidrule(r){1-4}
    PD-E & \textbf{41.9} & 31.4 & 19.6 \\
    \bottomrule
  \end{tabular}
\end{table}

During training, we observed that T-base and PD-H have difficulties to find attention alignment when the amount of data for fine-tuning is small. To be specific, T-base and PD-H failed to find attention for 0.4-hour fine-tune data, and only T-base failed when 2-hour fine-tune data were used. On the other hand, PA-HL and PD-E were successfully trained in all cases. From the result of Table \ref{table-preference-bli}, we can see that the raters preferred PA-HL in every competing pair. The proposed approach, PA-HL, could utilize information of the high-resource language for learning the low-resource language, and this led the model to show better generation quality. The WER of each approach in Table \ref{table-wer} shows a similar tendency with the preference test. We think the value of WER is higher than it sounds, probably because the speech recognition model had not been exposed to a synthesized speech during training. Still, we can compare the relative performance between models from the WERs.

\subsection{Multilingual pre-training for other languages}
\label{exp-css}

We validated our approach by applying the same pre-training framework in other languages. We repeated the same experiment which is done in Subsection \ref{exp-pretrain} with datasets different language. For each language, we used 2 hours of data to simulate low-resource language and compared PA-HL to PD-H by preference tests. Table \ref{table-preference-css} shows the result of the preference test of each language. While the performance gaps varied from language to language, PA-HL outperformed in most of the languages as it was in the previous subsection. We conjecture that the varying performance gap came from the difference in the phonology of each language, and we will investigate it in the future works.

\begin{table}[t]
  \caption{Result of preference test based on a 7-point rating scale for models trained with CSS dataset.}
  \label{table-preference-css}
  \centering
  \begin{tabular}{ccccc}
    \toprule
    \multirow{3}{*}{\shortstack[c]{Language}} & \multicolumn{3}{c}{Preference ($\%$)} \\
    \cmidrule(r){2-4}
     & PD-H & Neutral & PA-HL \\
    \cmidrule(r){1-4}
    Chinese & 25.0\% & 36.0\% & \textbf{39.0\%} \\
    Dutch & 28.0\% & 20.0\% & \textbf{52.0\%} \\
    Finnish & 30.0\% & 26.7\% & \textbf{43.3\%} \\
    French & 3.0\% & 6.0\% & \textbf{91.0\%} \\
    German & 14.0\% & 24.0\% & \textbf{62.0\%} \\
    Greek & 28.3\% & 13.3\% & \textbf{58.3\%} \\
    Hungarian & 25.6\% & \textbf{37.8\%} & 36.7\% \\
    Japanese & 25.0\% & 28.0\% & \textbf{47.0\%} \\
    Russian & 19.2\% & 20.8\% & \textbf{60.0\%} \\
    Spanish & 20.0\% & 20.0\% & \textbf{60.0\%} \\
    \bottomrule
  \end{tabular}
\end{table}

\section{Conclusion}
\label{conclusion}
In this work, we have trained multilingual multi-speaker TTS models using pairs of two monolingual datasets. By investigating the distance between learned phoneme embeddings, we have experimentally shown that the embeddings can represent the relation of pronunciations across the different languages. We proposed a pre-training framework that utilizes information of high-resource language to help training of low-resource language TTS model. The proposed approach outperformed other pre-training frameworks as well as the baseline Tacotron. Furthermore, by applying this pre-training framework to other 10 languages, we validated that the proposed method is generalizable to other pairs of languages.

Also, we are interested in applying the findings of this work to train an automatic speech recognition (ASR) system of low-resource languages. Like TTS, ASR also requires a large amount of training data. We may train a TTS model as proposed in this work. Since the TTS model can generate speech of arbitrary sentence in various speakers' voice, it will improve the robustness of the ASR model. We hope to investigate how the multilingual TTS can help training of the ASR model in the future.

\bibliographystyle{IEEEtran}

\bibliography{mybib}

\end{document}